\documentclass[twocolumn, switch]{article}
\usepackage{preprint}
\usepackage{hyperref}
\usepackage[numbers,square]{natbib}

\usepackage{amsmath}
\usepackage{booktabs}
\usepackage{graphicx}
\usepackage{glossaries}
\makeglossaries

\hypersetup{colorlinks=true, linkcolor=purple, urlcolor=blue, citecolor=cyan, anchorcolor=black}

\usepackage[utf8]{inputenc}	
\usepackage[T1]{fontenc}
\usepackage{xcolor}
\usepackage{lineno}					
\usepackage{tikz} 					

\usepackage{newfloat}
\DeclareFloatingEnvironment[name={Supplementary Figure}]{suppfigure}
\usepackage{sidecap}
\sidecaptionvpos{figure}{c}

\usepackage{floatrow}
\DeclareFloatFont{scriptsize}{\scriptsize}
\usepackage{titlesec}
\titlespacing\section{0pt}{12pt plus 3pt minus 3pt}{1pt plus 1pt minus 1pt}
\titlespacing\subsection{0pt}{10pt plus 3pt minus 3pt}{1pt plus 1pt minus 1pt}
\titlespacing\subsubsection{0pt}{8pt plus 3pt minus 3pt}{1pt plus 1pt minus 1pt}

\definecolor{lime}{HTML}{A6CE39}
\DeclareRobustCommand{\orcidicon}{
	\begin{tikzpicture}
	\draw[lime, fill=lime] (0,0)
	circle [radius=0.16]
	node[white] {{\fontfamily{qag}\selectfont \tiny ID}};
	\draw[white, fill=white] (-0.0625,0.095)
	circle [radius=0.007];
	\end{tikzpicture}
	\hspace{-2mm}
}

\title{Dynamic Gaussian Splatting from Markerless Motion Capture can Reconstruct Infants Movements}

\usepackage{authblk}

\author[1, 2\thanks{\texttt{rcotton@sralab.org}}]{R. James Cotton\href{https://orcid.org/0000-0001-5714-1400}{\orcidicon}}
\author[3]{Colleen Peyton}
\affil[1]{Shirley Ryan AbilityLab}
\affil[2]{Northwestern University, Department of Physical Medicine and Rehabilitation}
\affil[3]{Northwestern University, Department of Physical Therapy}

\begin{document}

\twocolumn[\begin{@twocolumnfalse}

\maketitle

\begin{abstract}
Easy access to precise 3D tracking of movement could benefit many aspects of rehabilitation. A challenge to achieving this goal is that while there are many datasets and pretrained algorithms for able-bodied adults, algorithms trained on these datasets often fail to generalize to clinical populations including people with disabilities, infants, and neonates.  Reliable movement analysis of infants and neonates is important as spontaneous movement behavior is an important indicator of neurological function and neurodevelopmental disability, which can help guide early interventions. We explored the application of dynamic Gaussian splatting to sparse markerless motion capture (MMC) data. Our approach leverages semantic segmentation masks to focus on the infant, significantly improving the initialization of the scene. Our results demonstrate the potential of this method in rendering novel views of scenes and tracking infant movements. This work paves the way for advanced movement analysis tools that can be applied to diverse clinical populations, with a particular emphasis on early detection in infants.\\
\end{abstract}

\keywords{rehabilitation, human pose estimation, infants, dynamic gaussians}

\vspace{0.5cm}

\end{@twocolumnfalse}]


\section{Introduction}

\begin{quote}
"What I cannot create, I do not understand"  - Richard Feynman
\end{quote}

There is a pressing need for high-quality movement analysis in rehabilitation and advances in computer vision and human pose estimation are moving closer to filling this gap. A common challenge in applying human pose estimation to rehabilitation populations is that algorithms trained on able-bodied adult populations fail to generalize to clinical populations \cite{cimorelli_portable_2022}. Tracking infants raises additional challenges, as they have different anthropomorphic body portions and movement dynamics than adults, and there is very limited training data.

Spontaneous infant movement behavior, that is endogenously generated, is an important indicator of neurological function and neurodevelopmental disability \citep{einspieler_qualitative_1997}. Cerebral palsy, the most common physical disability of childhood \citep{oskoui_update_2013}, can be predicted with high accuracy from clinical assessment of spontaneous movement behavior in infants \citep{bosanquet_systematic_2013}. This clinical assessment called the General Movement Assessment (GMA) \citep{einspieler_qualitative_1997}, is used by trained clinicians who use their gestalt perception to distinguish and identify various movement patterns in the young infant from the preterm period to 5 months corrected age. The high accuracy of this assessment highlights the importance of movement behavior in understanding the nervous system that generates it. On the other hand, there is a lack of objective and quantitative knowledge surrounding early movement generation in young infancy.

As the field of neonatology has advanced, infants born preterm are surviving at younger gestational ages, providing a unique opportunity to study the development of joint kinematics prior to term age. These preterm movement kinematics are likely to provide an early prognostic biomarker to identify infants at risk of neurodevelopmental disability, which could enable early intervention programs. However, this opportunity has not yet been realized as unobtrusive methods to reliably measure movement during the preterm period have not often been employed. During the preterm period, infants have fragile skin, small limbs, and sensitivity to touch which can cause autonomic dysregulation, prohibiting the use of traditional measurement approaches that require sensors or markers to be placed on the body.

Therefore, human pose estimation is a potential solution to study infant movements at younger ages. Several studies have used this approach and trained infant movements to predict clinical assessments such as the GMA. For example, using a small public dataset (n=12), a complexity score of infant limb movement was created to predict GMA rating \citep{wu_rgb-d_2021}, which was then tested prospectively on a larger cohort of 47 infants with high specificity for detecting a normal GMA finding \citep{wu_automatically_2021}. As an alternative to the GMA, the Computer-based Infant Movement Assessment (CIMA) model performs a time-frequency decomposition of movements estimated from video with both manual annotation and optical flow from 3-month-old infant data to predict a diagnosis of cerebral palsy at $\ge$ 2 years of age in a sample of 377 infants with comparable sensitivity and specificity to the GMA \citep{ihlen_machine_2019}.

Acquiring high-quality training data on infant movements is also challenging. The markers used for marker-based motion capture are quite large compared to babies and will often be knocked off. Similarly, most wearable sensors are quite large compared to infants and neonates and are time-consuming to place, making them difficult to use in clinical practice, particularly for premature babies in the neonatal intensive care unit (NICU). There are a few existing small datasets including the MINI-RGBD dataset \citep{hesse_computer_2019} which contains 12 movement sequences computed with synthetic textures on the Skinned-Multi Infant Linear Model (SMIL) \citep{frangi_learning_2018} model to preserve privacy. This model was learned from data acquired with RGB and depth imaging from a Kinect sensor. Clinicians assessing the GMA from SMIL reconstructions of movement from RBG-D images showed moderate-good agreement for movement complexity and substantial-good agreement for fidgety movements, compared to scoring the real videos.  There is also the SyRIP dataset, which contains a small sample of synthetic and real infants \citep{huang_invariant_2021}. \citet{groos_towards_2022} collected a diverse dataset of 20k frames, including in-hospital and at-home video, and found that an EfficientHourglass architecture trained on this dataset showed only a slightly greater error than the spread between human annotators.

Advances in 3D scene representations allow scenes to be learned from a variety of data and allow rendering of novel views. This includes methods using implicit representations of the volumes \citep{gao_nerf_2022}. This is most often applied to static scenes from a large number of images taken in different positions, with the camera calibration identified as an initial step. Very recently, a similar visual performance at much faster rendering rates was achieved by replacing the implicit representation with a set of 3D Gaussians \citep{kerbl_3d_2023}. This was enabled through a high-performance differential renderer that performs splatting of the Gaussians to camera views, allowing the parameters of Gaussians to be optimized through gradient descent. This approach was then extended to dynamic scenes \citep{luiten_dynamic_2023} that were recorded with 27 RGB cameras and additional depth cameras, by allowing the Gaussians to move over time. In addition, to enable synthesizing novel views of the subject in the scene, the Gaussians tracked meaningful body parts. Thus this approach is promising for movement analysis, both to track movements of all body parts and for synthesizing novel views of the scene that could serve as training data. We note in the last month several other works on dynamic Gaussians, including using 4D Gaussian representation that includes a time component \citep{wu_4d_2023, yang_real-time_2023} and using an implicit function to model the temporal deformations of the points to reconstruct scenes from monocular views using structure from motion to initialize the scene \citep{yang_deformable_2023}. Older work has also show that a similar differentiable rendering of 72 gaussians can be coupled to an anatomical model over configurations to perform human pose estimation \citep{rhodin_versatile_2016}.

The goal of this work was to test whether dynamic Gaussian tracking could be applied to a sparse set of RGB cameras, a configuration commonly used for markerless motion capture (MMC) \citep{cotton_markerless_2023, uhlrich_opencap_2022, kanko_assessment_2021}. MMC is seeing increasing use in rehabilitation, with applications to date predominantly focusing on gait analysis. However, achieving accurate scene reconstruction and novel field synthesis from MMC data could enable a wider range of applications.  However, this is challenging, because the Gaussians locations are typically initialized using either very dense image capture or additional depth imaging. With only sparse RGB images, the initial scene reconstruction is underconstrained. We show that using a semantic segmentation mask to only reconstruct the infant results in substantial improvements in the reconstruction of unseen views. With this improved initialization, we find the dynamic Gaussians can pick up spontaneous movements of the babies and visualize this movement from novel views.

In short, our contributions are:

\begin{itemize}
\item We apply dynamic gaussian splatting to sparse, markerless motion capture data
\item We optimize this reconstruction method and perform ablations to demonstrate the importance of semantic information and masking
\item Show this allows us to track 3D movement of babies
\end{itemize}

\section{Methods}

\subsection{Participants}


This study was approved by our Institutional Review Board. Two infants born at term age were recruited. One infant was filmed at 3 weeks of age and again at 5 weeks of age. The second infant was filmed at 12 weeks of age. Synchronized videos were recorded as the infants were positioned in a calm, alert state, without a pacifier, for short periods on the mats, in order to observe spontaneous behavior.

\subsection{Video acquisition}

Multicamera data was collected with a custom system using 8 FLIR BlackFly S GigE cameras with F1.4/6mm lens. They were synchronized using the IEEE1558 protocol and acquired data at 30 fps, with a typical spread between timestamps of less than 100µs. The images produced have a width of 2048 and a height of 1536. The cameras were mounted on tripods which were placed in a circle around a padded mat on the floor. The acquisition software was implemented in Python using the PySpin interface. For each experiment, calibration videos were acquired with a checkerboard ($7 \times 5$ grid of 38mm squares). Extrinsic and intrinsic calibration was performed using the anipose library \citep{karashchuk_anipose_2020}. The intrinsic calibration included only the first distortion parameter.

\subsection{3D Gaussian Splatting}

Our method is built upon \citet{luiten_dynamic_2023}, which is a dynamic extension of 3D Gaussian Splatting \citep{kerbl_3d_2023}. Gaussian splatting directly optimizes the parameters of a set of 3D Gaussian kernels to reconstruct a scene observed from multiple cameras. It is powered by a custom CUDA kernel for fast, differential rastering engine. Each Gaussian is described by opacity, color, location, scale, and rotation.

The opacity, $\alpha$, is a scalar value and the color, $c$, is a 3-vector, both of which are between 0 and 1 after passing through a sigmoid transformation. The location, $\mu$, is a 3-vector for the center of the Gaussian in Euclidean space, where we used meters as the units. The scale is a 3-vector describing the spatial extent of the Gaussian in each dimension and has an exponential non-linearity to ensure positivity. The rotation is a quaternion, which includes a non-linearity to ensure it is normalized to have a unit norm.

The potential influence of each gaussian on any location in space is computed from:

\begin{equation}
G(\mathbf{x}) = \sigma(o) \, e^{-\frac{1}{2} (\mathbf{x}-\mathbf{\mu})^T \mathbf{\Sigma}^{-1} (\mathbf{x}-\mathbf{\mu})}
\end{equation}

Where the spatial covariance is determined by the spatial scale, $S \in \mathbb{R}^3$, and the rotation, $R \in \mathbb{R}^{3 \times 3}$, which is computed from the quaternion representation:

\begin{equation}
\Sigma = RSS^\top R^\top
\end{equation}

Combined with the opacity and color, the ensemble of Gaussians is efficiently ray-traced with the differentiable renderer. Specifically the color of each pixel is computed as:

\begin{equation}
C = \sum_{i \in N} T_i \alpha_i c_i
\end{equation}

Where the transmittance, $T_i =\prod_{j=1}^{i -1} \alpha_j$ is computed based on the opacities of the Gaussians traced along the ray.

This takes in the intrinsic (without any distortion coefficients) and extrinsic parameters of the calibrated cameras. We refer to \citep{kerbl_3d_2023} for further details about the rendering engine, which includes many features for depth-sorting and spatially culling Gaussians into patches to allow it to quickly render millions of elements and also includes explicit deviations of the derivatives in the handwritten CUDA kernels.

Following \citet{kerbl_3d_2023, luiten_dynamic_2023} a loss is computed between the reconstructed images and the observed images that includes both an L1 term D-SSIM term, where we also use a relative weighting of $\lambda=0.2$.

\begin{equation}
\mathcal{L}_{\mathrm im} = (1 - \lambda) \mathcal{L}_1 + \lambda \mathcal{L}_{\text{D-SSIM}}
\end{equation}

\citet{luiten_dynamic_2023} extended this approach to include an additional color component for each Gaussian, which corresponds to a segmentation map between foreground and background, although can flexibly correspond to any secondary color information.  This uses the same loss function as the regular image, $\mathcal L_\mathrm{seg}$. We discuss the segmentation mask further below. \citet{luiten_dynamic_2023} also replaced view-dependent spherical harmonic representations of color for having isotropic colors with learnable scales and means for each camera, and we retained this feature.

Because the differentiable renderer does not account for camera distortions, we applied the OpenCV \citep{opencv_library} \texttt{undistort} method to our raw images.

\subsection{Dynamic Losses}

\citet{luiten_dynamic_2023} includes several additional losses for the Gaussians between timesteps.  Each of these is applied to a local region of Gaussians identified using KNN clustering based on the initial scene reconstruction.

\begin{equation}
\mathcal{L}^{\text{rigid}}_{i,j} = w_{i,j} \left\| (\mu_{j,t-1} - \mu_{i,t-1}) - R_{i,t}^{-1} R_{i,t} (\mu_{j,t} - \mu_{i,t}) \right\|_2
\end{equation}

\begin{equation}
\mathcal{L}_{\text{rigid}} = \frac{1}{k|S|} \sum_{i \in S} \sum_{j \in \text{knn}_i,k} \mathcal L^{\text{rigid}}_{i,j}
\end{equation}

\begin{equation}
\mathcal{L}_{\text{rot}} = \frac{1}{k|S|} \sum_{i \in S} \sum_{j \in \text{knn}_i,k} w_{i,j} \left\| \hat{q}_{i,j,t}^{-1} - \hat{q}_{i,t}\hat{q}_{i,t-1} \right\|_2
\end{equation}

\begin{equation}
\mathcal{L}_{\text{iso}} = \frac{1}{k|S|} \sum_{i \in S} \sum_{j \in \text{knn}_i,k} w_{i,j} \left( \left\| \mu_{j,0} - \mu_{i,0} \right\|_2 - \left\| \mu_{j,t} - \mu_{i,t} \right\|_2 \right)
\end{equation}

\subsection{Optimization}

%
%

We followed prior work and used the Adam optimizer with different learning rates for each of the parameters. The learning rate for the mean Gaussian locations for 0.00016 times the scale of the scene. The colors had a learning rate of 0.0025. The segmentation map had a learning rate of 0.001. The unnormalized quaternion representation had a learning rate of 0.001. The logit for the opacities had a learning rate of 0.05. The logarithm of the scales had a learning rate of 0.001. The camera means and scales (pre-exponential) had a learning rate of 1e-4.

The optimization process also includes additional heuristics during training that use the accumulated derivates applied to each Gaussian to determine regions that should either have be pruned or have the density increased. Points with very low opacities are also pruned, as are points with spatial scales that exceed a threshold size (typically 10\% of the scene volume). We refer to the prior work for these details. For dynamic fits, parameters from the previous frame were used to initialize the next frames, as in \citep{luiten_dynamic_2023}, with the velocity over the prior two frames used to predictively adjust the Gaussian positions prior to the optimization. We used 2000 iterations of optimization for subsequent frames.

We rendered the scenes at the full 2048x1536 resolution, which ran at approximate 20 iterations per second on an A6000 (with two renderings per iteration to produce the segmentation mask). Similarly to \citep{luiten_dynamic_2023, kerbl_3d_2023} we sampled the training views in a random order in blocks without replacement.

\subsection{Initialization}

In \citet{kerbl_3d_2023, luiten_dynamic_2023}, the Gaussians are initialized from a precomputed point cloud. In the case of \citet{luiten_dynamic_2023}, which performed dynamic reconstruction with human movement recorded with 27 cameras, this point cloud was initialized with an additional set of depth cameras. We found that with random initialization, optimization of the first frame from a random point cloud did not converge to an accurate reconstruction of the scene, as reflected both by poor generalization to camera views not used during reconstruction and by visually inspecting the optimized point cloud.

In particular, we noted a lot of background Gaussians obscuring the validation view and initial experiments adjusting the pruning and density-increasing heuristics did not seem to resolve this. Visualizing interpolation between camera views made it apparent how pieces of floating geometry would align into specific places for the training views to reconstruct the correct images with an incorrect geometry, highlighting the challenges of reconstruction from these underconstrained data.

To improve convergence to a reconstruction that matches the underlying geometry, we explored using additional visual cues to guide the scene reconstruction.

\subsection{Depth}

We attempted to provide additional supervision from inferred depth maps to remove the sparse, noisy, background Gaussians. This was motivated by recent work showing that depth was a necessary supervisory signal when using implicit representations to reconstruct dynamic scenes from sparse RGB-D images \citep{gerats_dynamic_2023}. Because our cameras only produce RBG data, we used DistDepth \citep{wu_toward_2022} to infer the depth images, which produced plausible results. The official CUDA implementation of the differentiable renderer does not support backpropagation through the depth image, so we used a fork that implemented this functionality \citep{yang_ingra14mdepth-diff-gaussian-rasterization_2023}. We followed \citet{luiten_dynamic_2023} by including a learnable per-camera scale and offset when comparing the rendered depth maps to the inferred maps, using an L1 loss. However, even with this additional supervision, we were unable to achieve an accurate inital reconstruction of the underlying geometry. We found comparing the depth images to the rendered depth images was still a useful diagnostic tool, and include these images in our results.

\subsection{Segmentation and Masking}

We then tried to use semantic segmentation to provide additional supervision. We used the Mask2Former implementation available through the Huggingfaces library \citep{cheng_masked-attention_2022}. Specifically, we used the large model with a Swin background, trained on the ADE20K dataset, which outputs 150 class labels \citep{liu_swin_2021, zhou_scene_2017}. We mapped these 150 classes to different RGB colors, with a black background to produce the segmentation target image for each view. We did include code to map the \texttt{plaything, toy} class to the \texttt{person} class as we saw a few instances where the infant was misclassified as a toy (perhaps reflecting dolls in the training dataset).

In addition to providing an additional semantic segmentation image, which was used to compute the segmentation loss, $\mathcal L_{seg}$, we also supported masking the image to only include the infant. This was done by first identifying the largest contiguous mask, and only including this (the legs of experimenters and parents were sometimes visible). Then the raw image was set to zero outside the identified mask, and this masked image was used to compute the image loss, $\mathcal L_\mathrm{im}$. When masking, we also added the ability to prune Gaussians outside the volume where the infant was placed (within 2 meters of the calibration center near the mat).

\subsection{Sparsity pruning}

Even with masking during the initial reconstruction, we still found some Gaussian scattered through open areas that would obscure novel views. During the densification and pruning stages already implemented, we added an addition pruning step that would remove these points. Specifically, we would remove any Gaussian that had a minimum distance to their nearest other Gaussian that was greater than 0.1m. Like this other steps of density adjustment, this occurred for every 100th iteration between 500 and 15000. In initial experiments, we attempted making this a differentiable term in the loss function, but found that computing the pairwise distances was very slow and still was not producing the desired results.

%
%
%

\subsection{Training losses}

This gave us a total loss for initialization of:

\begin{equation}
\mathcal L_{\text{init}} = \lambda_\mathrm{im} \mathcal L_\mathrm{im} + \lambda_\mathrm{seg} \mathcal L_\mathrm{seg}
\end{equation}

And for the subsequent frames of

\begin{equation}
\mathcal L_{\text{dyn}} = \lambda_\mathrm{im} \mathcal L_\mathrm{im} + \lambda_\mathrm{seg} \mathcal L_\mathrm{seg} + \lambda_\mathrm{rigid} \mathcal L_\mathrm{rigid} + \lambda_\mathrm{rot} \mathcal L_\mathrm{rot} + \lambda_\mathrm{iso} \mathcal L_\mathrm{iso}
\end{equation}

With parameters $\lambda_\mathrm{im}=1.0$, $\lambda_\mathrm{seg}=3.0$, $\lambda_\mathrm{rigid}=4.0$, $\lambda_\mathrm{rot}=4.0$, and $\lambda_\mathrm{iso}=2.0$.

\subsection{Metrics}

To quantify the performance of the reconstruction, we used 7 cameras to reconstruct the scene and measured the accuracy of the reconstructed images against this validation view. The PSNR is computed from the L2 loss and is a measure of the peak signal-to-noise ratio. In all cases, this was only applied to the region of the image with the segmentation mask corresponding to the baby.

\begin{equation}
\text{PSNR} = 20 \log_{10} \left( \frac{1}{\sqrt{\text{MSE}}} \right)
\end{equation}

\subsection{Optimizating initial scene reconstruction}

We explored a range of parameters to determine their impact on the initial scene reconstruction, as this was essential to good dynamic performance. To quantify this we took a set of frames from two sessions, one from each baby. For each scene, we repeated the reconstruction using one of four cameras as the validation cameras and used the remaining 7 cameras to perform the scene initial reconstruction. For each of the 8 fits, we computed the metrics described above for the validation frame.

\subsection{Dynamics through deformation fields}

We also implemented an alternative dynamic tracking method using deformation fields, based on \citet{yang_deformable_2023}. Instead of iteratively updating the location of Gaussians for each frame with a regularizer based on the prior frame location, we trained a deformation field that models the change in location, rotation and scale based on the initial 3D locations and the desired time point. This was implemented as:

\begin{equation}
(\delta \mu, \delta r, \delta s) = F_\theta (\gamma(x), \gamma(y), \gamma(z), \gamma(t))
\end{equation}

\begin{equation}
\gamma(p) = \left( \sin(2\pi p), \cos(2\pi p) \right)^{L-1}_{k=0}
\end{equation}

where $\gamma(p)$ is sinusoidal positional encoding applied to each of space-time coordinates, with $L=10$ and $F_\theta$ is an 8 layer MLP with hidden dimension of 512.

We performed optimization of the initial scene as above for 9000 iterations, after which we performed iterations of the entire sequence for another 40k iterations. When optimizing the entire sequence, we only used the loss for the image $\mathcal{L}_{\mathrm im}$. Images from the training views were randomly sampled from different timepoints. Because this required entire sequence to be loaded into memory, the sequence length is limited by system memory. As in \citet{wu_4d_2023}, we added Gaussian noise to the positional encoded value of time with a standard deviation of 0.1 that linearly reduced to 0 after 20k iterations. The found this served to anneal the solution with training and improve the temporal performannce.

Because the deformation field does not not use an an explicit map onto the Gaussians, but rather takes in their coordinates, it was possible to continue performing the density adjustments while training the entire sequence. We kept this component unaltered from above.

\section{Results}

\subsection{Initial static reconstructions}

First we tested the impact of our additional supervision sources on the initial scene reconstruction, showing that using the mask from the semantic segmentation was critical.

\begin{figure}[!htbp]
\centering
\includegraphics[width=1\linewidth]{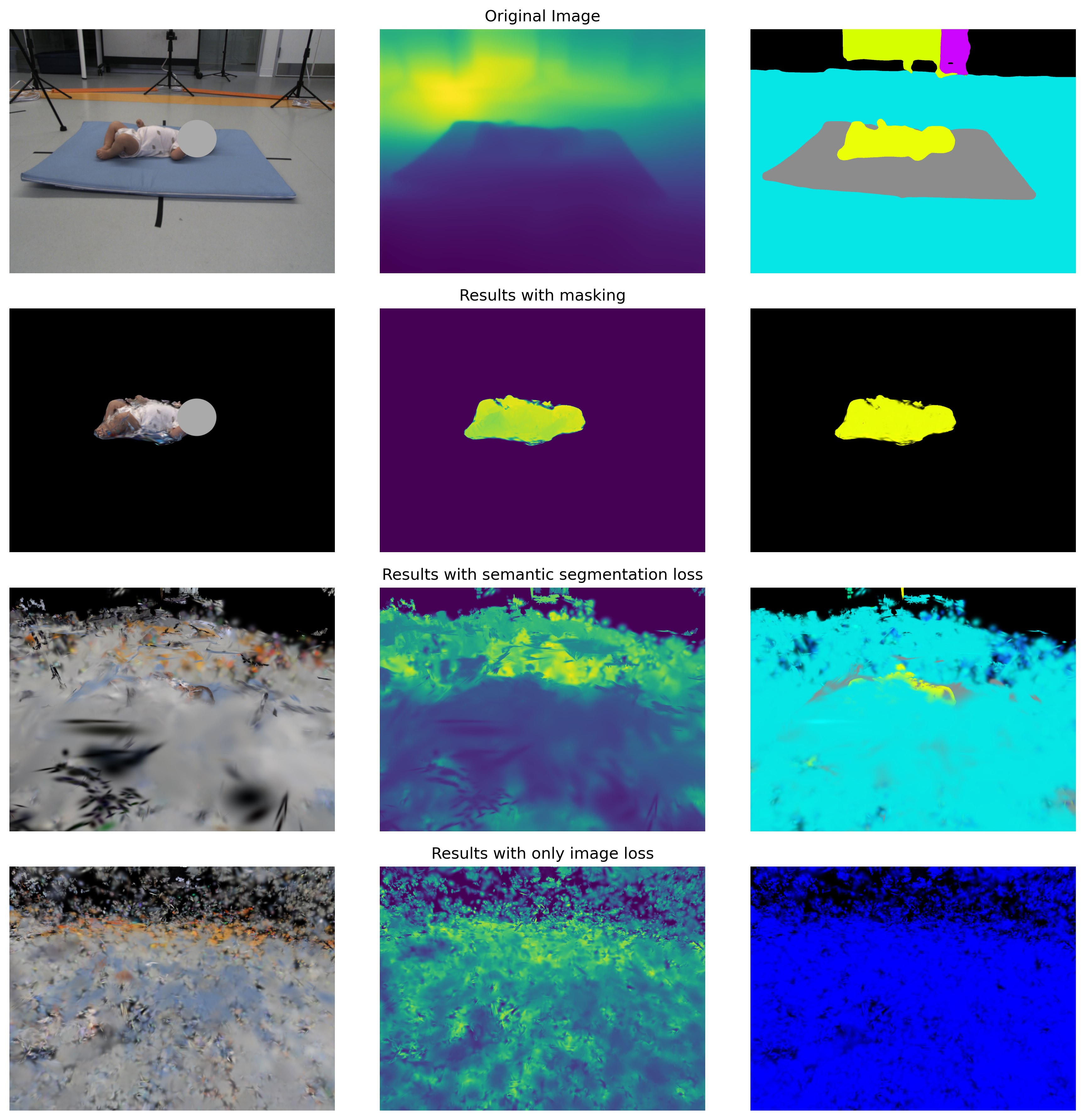}
\caption[]{Example visualizations of the validation view reconstructed with the different supervision signals. The top row shows the original image with the depth image and semantic segmentation mask (the later two both inferred with algorithms described above). The next row shows the image, depth image, and semantic map reconstructed when the baby mask was applied to all views. The next row uses the semantic segmentation as a supervision signal, but is unable to reconstruct novel views other than hints of the semantic map. The final row shows that with no additional supervision, the reconstruction is very poor.}
\label{ablations}
\end{figure}

\begin{figure}[!htbp]
\centering
\includegraphics[width=1\linewidth]{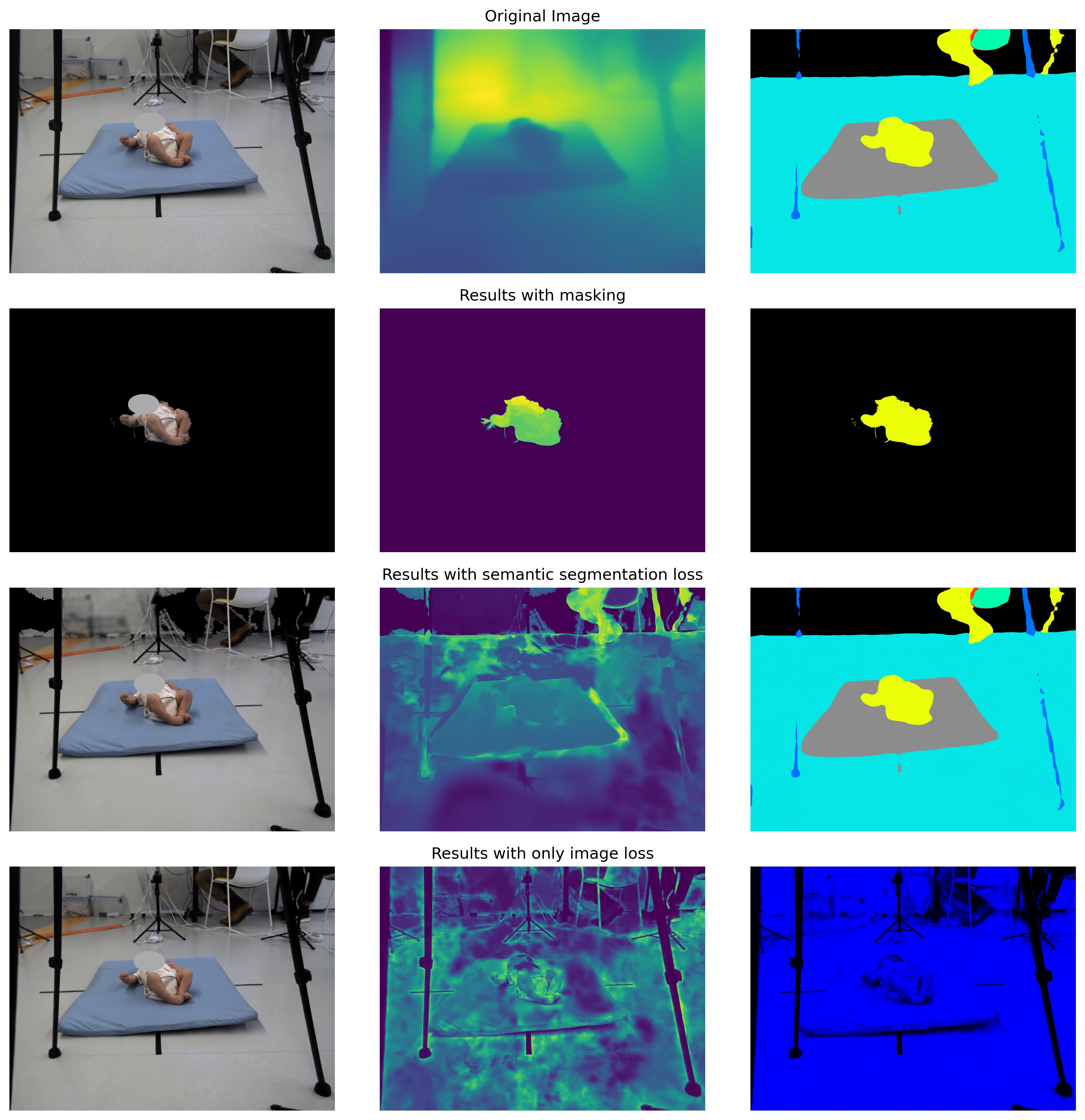}
\caption[]{The same as Figure~\ref{ablations}, but for the training views. In all settings, Gaussians are optimized that can reconstruct the training views. However, the depth maps already show inconsistencies reflecting the floating geometry aligned to the camera perspective. Note that on the third row, the model is able to recreate the segmentation map.}
\label{ablations_training}
\end{figure}


\begin{table}
\centering
\caption[]{Settings and results of ablation studies showing dramati decline in validation PSNR when not including the segmentation mask.}
\label{ablation-table}
\begin{tabular}{p{\dimexpr 0.143\linewidth-2\tabcolsep}p{\dimexpr 0.143\linewidth-2\tabcolsep}p{\dimexpr 0.143\linewidth-2\tabcolsep}p{\dimexpr 0.143\linewidth-2\tabcolsep}p{\dimexpr 0.143\linewidth-2\tabcolsep}p{\dimexpr 0.143\linewidth-2\tabcolsep}p{\dimexpr 0.143\linewidth-2\tabcolsep}}
\toprule
Mask & $\lambda_\mathrm{seg}$ & Spare & \# Init Points & Init Range & Val PSNR & Train PSNR \\
\hline
True & 3.0 & False & 10000 & 1.5 & 13.24 & 21.09 \\
True & 3.0 & False & 20000 & 1.5 & 13.31 & 20.94 \\
False & 3.0 & False & 100000 & 10.0 & 7.79 & 25.94 \\
False & 0.0 & False & 100000 & 10.0 & 7.96 & 34.29 \\
True & 3.0 & True & 10000 & 1.5 & \textbf{13.41} & 20.71 \\
\bottomrule
\end{tabular}
\end{table}

\subsubsection{Novel view synthesis with masking}

With masking enabled and discarding all Gaussians outside 2m from the center of the mat, we got visually compelling results for novel views Figure~\ref{ablations} (second row). The PSNR was for the validation view (computed over the region of the baby mask, only) averaged over our 8 training conditions was about half of the PSNR for the training views Table~\ref{ablation-table}.

\subsubsection{Ablations}


We originally attempted to reconstruct without additional constraints. In this case, we did not prune any Gaussians based on their spatial location and initialized 100k points uniformly over a 10m area (instead of 10k for the smaller area). We found that while this was able to match the training views, the validation views were heavily obscured by random floating Gaussians Figure~\ref{ablations} (fourth row). This was true even when reconstruction of the training images performed well Figure~\ref{ablations_training} (fourth row).

We also attempted initialization with the semantic segmentation loss. However, the views were still very obscured Figure~\ref{ablations} (third row). Note however that in this case the segmentation map was also reconstructed well for training views Figure~\ref{ablations_training} (third row), with some hints showing through on the validation views.

Additionally, we noted the rendered depth images showed a clear discrepancy between the depth maps estimated with DistDepth, but as noted were unable to use this as a supervision signal while the depth channel is not backpropagated through the differential renderers CUDA kernel.

\subsubsection{Hyperparameter tests}

For the masked reconstruction, we also explored several other hyperparameter values. For example, we found that increasing the iterations from 6000 to 16000 steps did not substantially improve the performance and by 30000 steps showed signs of overfitting. After reducing to 2000 iterations, performance degraded. For most other adjustments, we used 6000 iterations to reduce the computation time. Within the range of parameters we adjusted, their impact was relatively small compared to the impact of the masking.

\subsection{Dynamic reconstructions}

\begin{figure}[!htbp]
\centering
\includegraphics[width=1\linewidth]{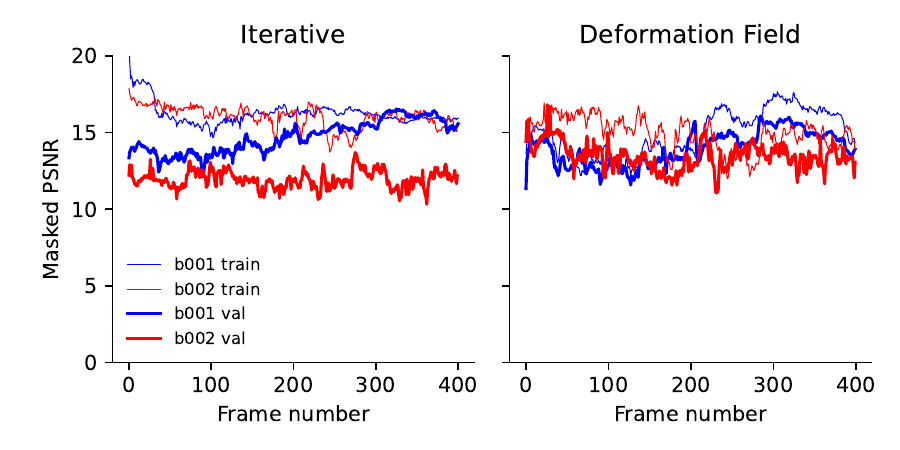}
\caption[]{Masked PSNR over dynamic tracking. The left column shows the training and validation PSNR for two videos fit with the iterative method. The right column shows the same but for the deformation field method.}
\label{dynamic_losses}
\end{figure}

Based on these results, we tested the performance of dynamic tracking with both the iterative tracking approach and the deformation field approach. Quantitatively, we saw that both approaches produced comparable PSNR values for the training data (Figure~\ref{dynamic_losses}). The iterative approach achieved higher PSNR values on the training views. Qualitatively, it appeared to produce sharper images in general. However, we also saw some Gaussians would drift away from the infant in the dynamic reconstruction, which create floaters in the rendered views. In a few frames, these would also create substantial visual artifacts. In contrast, the deformation field lost some of this higher resolution detail but more reliably captured the movements. Loading the entire sequence into memory used 300GB of system memory for 400 frames.  The fitting was also substantially faster than the iterative approach, with the 40k iterations taking about 40 minutes for 400 frames, compared to nearly 12 hours for this many frames updated iteratively with 2k steps per frame.

\begin{figure}[!htbp]
\centering
\includegraphics[width=1\linewidth]{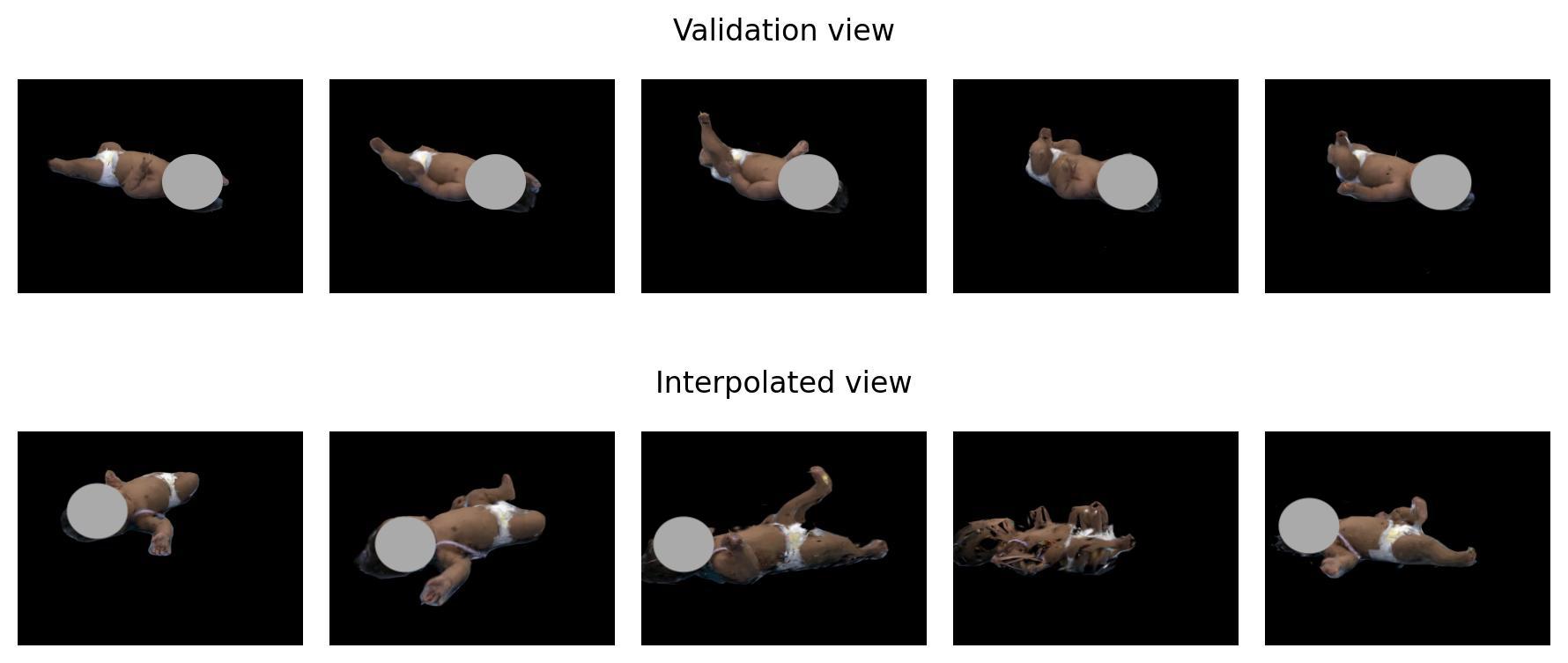}
\caption[]{Frames from iterative fit. The top row shows frames reconstructed from the validation cameras. The bottom row shows an interpolated camera view between two of the training views, with the middle three images not being from a training view. Timesteps were selected to capture different body postures. Notice the fourth interpolated frame shows substantial floaters obscuring the view that drifted during dynamic fitting.}
\label{dynamic_reconstruction}
\end{figure}

\subsection{Tracking}

\begin{figure}[!htbp]
\centering
\includegraphics[width=1\linewidth]{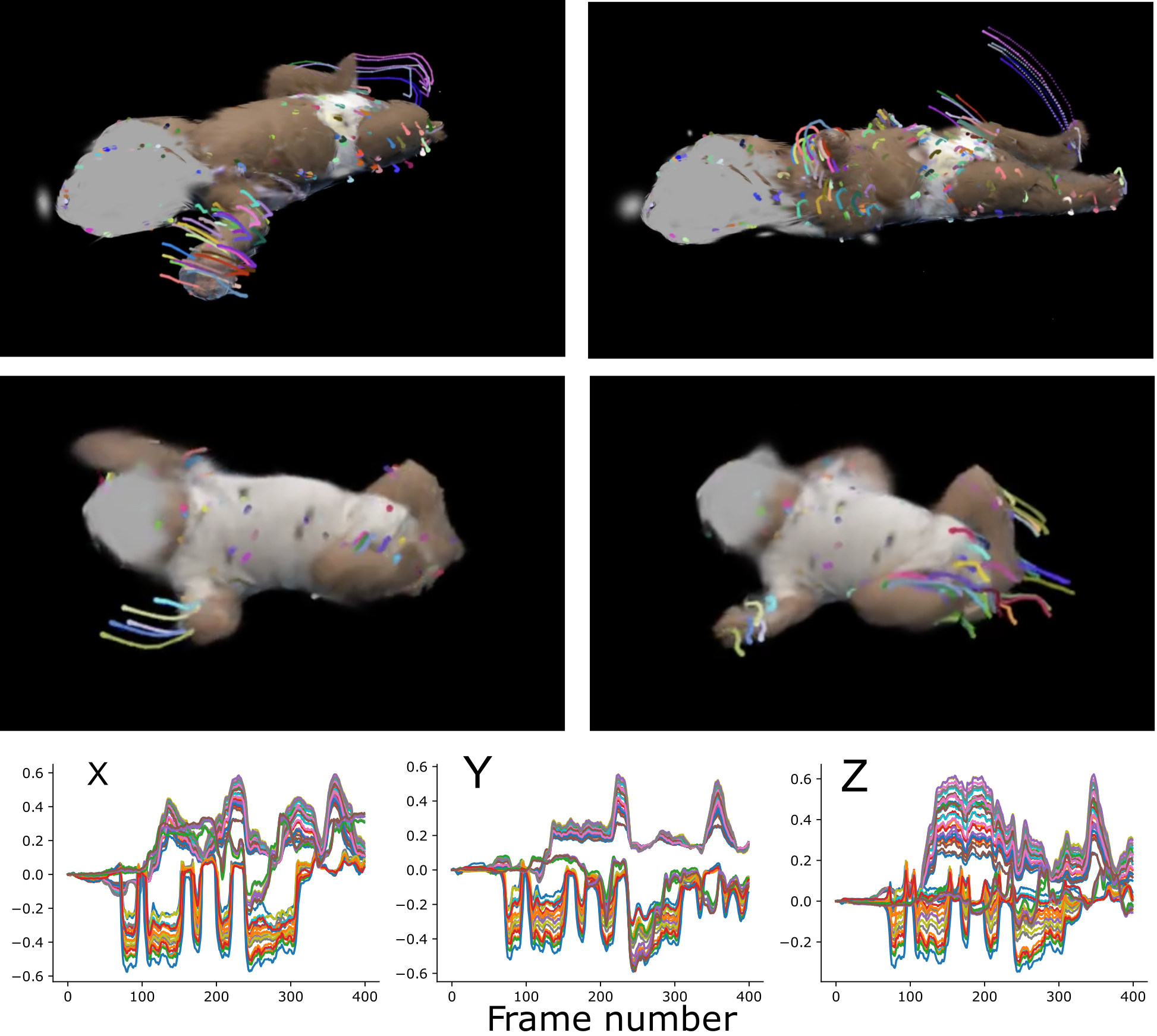}
\caption[]{Snapshots of movement with 2\% of Gaussians showing a trajectory of their history, with each point assigned a unique color. This shows how individual Gaussians track specific anatomic locations. The top row used the iterative dynamic tracking and the middle row used the deformation field. The bottom traces show a subset of displacements for Gaussians, showing groupings of coherent movement.}
\label{baby_tracking}
\end{figure}

We wanted to determine whether the Gaussians track specific body parts of the infant, as was seen in \citet{luiten_dynamic_2023}. We created videos that show the location history of 2\% of the Gaussians over the last 10 frames (interpolated up to 10x temporal resolution). We observed that as expected, individual color tracks persisted in alignment with specific body locations using both dynamic methods. For example Figure~\ref{baby_tracking} shows consistent traces over the leg during these movements, as well as movement over the arm, with the markers over the head showing much less movement. We did not quantify this consistency against any existing pretrained keypoint detectors.

\section{Discussion}

We find that dynamic Gaussian splatting to reconstruct markerless motion capture data shows promise for rendering novel views of scenes and tracking movements within the scene. However, a good initialization is critical and challenging. With only sparse views, many unrealistic geometries can reconstruct the training views. This becomes apparent when interpolating between views, as artifactual floating geometries obscure the interpolated views and then snap into place when viewed from the training image perspectives.

Applying a segmentation mask and only reconstructing the infant in the MMC view drastically reduces this problem and improves the realism of novel views. However, artifacts were still visible. Some were due to imperfect segmentation masks, causing Gaussians that capture the mat to attach to the infant. Sparsifying the geometry to remove any Gaussians that are more than 10cm from their nearest neighbor lessened these floating artifacts and improved the validation PSNR. In dynamic scenes, some Gaussians also drifted away from the infant. This suggests future opportunities to improve the priors and heuristics during initialization and on the dynamics between frames.

We expected that combining segmentation masks and depth as additional supervision signals would allow us to optimize realistic geometries for the entire scene. However, even combined with additional sparsification, the result was a cluttered reconstruction that generalized poorly to new views. It is possible even with the per-camera scale and offset, the depth images were not geometrically consistent, and we saw some artifacts in the inferred depth images that would suggest this such as offsets when the leg of a tripod passed through a view. We suspect there are other opportunities to use structure-from-motion algorithms to obtain a better initialization that would improve the whole-view reconstruction, and that this will also improve the reconstruction of the subject.

Iteratively updating the Gaussian locations versus using a deformation field seem to have different strengths. The iterative solution often produced sharper reconstructions, but was more prone to artifacts. This perhaps could refined through tuning the regularization losses between frames. The deformation field allows directly optimizing the entire dynamic scene, potentially allowing visual information from later in the sequence to improve earlier Gausisan locations. For example, the distribution of Gaussians in the initial scene over the surface seemed better distributed when optimizing with the deformation field. It also and takes less time to train. The memory limitations currently limit the sequence length that can be optimized, but a more efficient pipeling or caching to disk could remove this limitation. Because the deformation field was decoupled from the specific set of Gaussians, it was also possible to continue adjusting the density over the entire sequence. It also provides a natural framework to interpolate between frames and to compute the velocity of areas by differentiating with respect to time, which could potentially allow accounting for motion blurred and tracking faster movements.

In this work, we focused on infants. Our masking approach is particularly effective in this situation, where only a single infant is in view. However, we anticipate this approach will be extensible to room-scale tracking of all movements, allowing us to collect diverse training data to generalize to a wide clinical population. We anticipate this can be further enhanced by using a traditional approach with many images to create an initial set of Gaussians for the background, allowing full scene tracking.

Reconstructing the geometry underlying a scene to render novel views shows the amount of information extracted from a scene, but is not the most clinically useful representation of movement. This system could be further augmented with keypoint detectors or other algorithms to explicitly track semantically meaningful body components. In future work, we also anticipate using keypoint detectors to quantify the automatic tracking accuracy, as in \citep{luiten_dynamic_2023}. Additionally, we could fit the SMIL model \citep{frangi_learning_2018} to these points to produce a parametric representation that is comparable between participants. Ultimately, we hope to fit biomechanically valid models to this movement to provide a representation comparable to the language of clinicians and biomechanists.

We anticipate using this approach to synthetically render novel views from MMC datasets from a larger cohort of infants will allow us to train 2D and 3D keypoint detectors that generalize well to a wide range of infants, including those in the NICU.  If this enables algorithms that can predict clinical measurements of movement quality in infants, this will create a powerful approach to early detection and interventions for infants at risk for neurodevelopmental problems.

\section{Conclusion}

Dynamic gaussians can be used to reconstruct the underlying time-varying geometry of scenes from data collected with synchronous videos from only sparse views, such as from markerless motion capture systems. This allows rendering novel views and tracking the underlying geometries. This is a promising approach to tracking kinematics for individuals where there is limited training data for traditional pose estimation algorithms, such as the movements of infants, and could also produce training data for training these pose estimation algorithms.
\printglossaries

\section*{Acknowledgements}
\footnotesize
This work was generously supported by the Research Accelerator Program of the Shirley Ryan AbilityLab and with funding from the Restore Center P2C (NIH P2CHD101913). We thank the participants and staff of the Shirley Ryan AbilityLab for their contributions to this work.
\normalsize

\bibliography{main.bib}

\end{document}